\documentclass[11pt]{article}

\usepackage[preprint]{acl}

\usepackage{times}
\usepackage{latexsym}
\usepackage[T1]{fontenc}
\usepackage[utf8]{inputenc}
\usepackage{microtype}
\usepackage{inconsolata}
\usepackage{graphicx}
\usepackage{booktabs}
\usepackage{amsmath,amssymb}
\usepackage{xcolor}
\usepackage{pifont}
\usepackage{multirow}
\usepackage{array}
\usepackage{enumitem}
\usepackage{xspace}
\usepackage{listings}
\usepackage{tabularx}
\usepackage{colortbl}
\usepackage{CJKutf8}
\usepackage[listings,breakable,skins]{tcolorbox}

\lstdefinestyle{prompt}{basicstyle=\ttfamily\scriptsize, columns=fullflexible, keepspaces=true,
  breaklines=true, breakatwhitespace=true, breakindent=1.2em,
  postbreak=\mbox{\textcolor{gray}{$\hookrightarrow$}\space},
  frame=single, rulecolor=\color{gray!60}, framesep=3pt, xleftmargin=3pt, xrightmargin=3pt,
  aboveskip=6pt, belowskip=6pt, showstringspaces=false}

\newcommand{\cmark}{\ding{51}}
\newcommand{\xmark}{\ding{55}}

\newcommand{\evbad}[1]{\textcolor{red!70!black}{#1}}
\newcommand{\evgood}[1]{\textcolor{blue!70!black}{#1}}
\newcommand{\method}{MedGraph\xspace}
\newcommand{\bench}{MedMemoryBench\xspace}
\newcommand{\benchcite}{\citep{wang2026medmemorybench}\xspace}

\title{\method: Maintaining Patient State with Graph Memory for Long-Term Medical Dialogue}

\author{
  \textbf{Donghua Zhang\textsuperscript{1$\dagger$}} \quad
  \textbf{Zizheng Wang\textsuperscript{2*}} \quad
  \textbf{Ze Dong\textsuperscript{1*}} \quad
  \textbf{Yiding Feng\textsuperscript{3*}} \quad
  \textbf{Siyu Qian\textsuperscript{4}}
  \\
  \textsuperscript{1}University of Michigan \quad
  \textsuperscript{2}Georgetown University
  \\
  \textsuperscript{3}Johns Hopkins University \quad
  \textsuperscript{4}Harvard University
}

\begin{document}
\maketitle
{\renewcommand{\thefootnote}{*}\footnotetext{Equal contribution.}}
{\renewcommand{\thefootnote}{$\dagger$}\footnotetext{Corresponding author: \texttt{donghua@umich.edu}}}

\begin{abstract}
LLM-based medical assistants that follow a patient across visits need a memory that keeps the patient's clinical state correct as it changes.
We examine the errors of five memory systems on longitudinal medical dialogue and find three recurring causes: the retrieved evidence gives an outdated state, includes another person's health facts, or lacks a safety fact the answer depends on.
We present \method, a graph memory built for these three cases.
Construction extracts dated facts from each session, attributes each fact to the person it describes, and links medication, diagnosis, and measurement records into chains that keep earlier states.
Retrieval ranks the patient's chains as units, reads each chain at the time the question asks about, and adds the original dialogue text; a patient profile lists every recorded safety fact with every question.
On \bench, \method has the highest overall accuracy with both a GPT and a Qwen backbone, 10.5 and 10.4 points above the strongest of five baselines, with efficient memory construction and answering in the GPT setting.
\end{abstract}

\section{Introduction}
\label{sec:intro}

LLM-based healthcare agents increasingly support care across
multiple visits, including chronic disease management and
medication guidance \benchcite.
They must track the patient's current medications and reported
allergies, since recommending a discontinued medication or
overlooking an allergy can lead to unsafe advice.
Reliable LLM agents therefore need long-term memory that keeps
patient records accurate and up to date across visits.

\begin{figure*}[t]
  \centering
  \includegraphics[width=\textwidth]{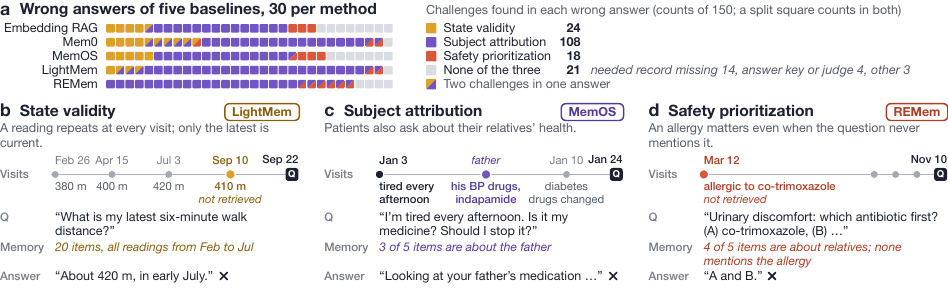}
  \caption{Three challenges in longitudinal medical memory.
  (a) We inspect 150 wrong answers from five memory systems,
  30 per system, and the evidence supplied to the answering
  model. Each square represents one wrong answer, with colors
  indicating the observed challenges. Split squares indicate
  two challenges; gray squares indicate failures outside these
  categories. Categories overlap, so their counts do not sum
  to 150.
  (b) LightMem misses the patient's September 10 walking-distance
  measurement, leading to an answer based on an older reading.
  (c) MemOS retrieves the father's medication when the patient
  asks about their own symptoms.
  (d) REMem omits a recorded co-trimoxazole allergy when answering
  a later question about antibiotics.}
  \label{fig:problem}
\end{figure*}
General-purpose LLM memory systems include retrieval-based
methods such as RAG \citep{lewis2020rag}, graph-based methods
such as GraphRAG \citep{edge2024graphrag}, and agentic approaches
such as Mem0 \citep{chhikara2025mem0}.
These systems differ in how they construct memory, including
what they retain and how they organize or update it, and how
they retrieve evidence for answering. For example, Mem0 extracts and updates conversational facts, while GraphRAG builds an entity-relation graph and community summaries. RAG retrieves similar passages, whereas GraphRAG uses graph-derived summaries.
Despite these advances, recent benchmarks report difficulties
in reasoning across medical dialogue sessions
\citep{hu2026medilongchat,zhang2026medlocomo}, and memory
accuracy declines as distracting sessions accumulate \benchcite.

To examine failures in memory construction and retrieval, we analyze errors from five representative memory systems.
This analysis reveals three challenges in longitudinal medical dialogue (Figure~\ref{fig:problem}).

\textbf{(1) State validity.} The same clinical item can have different values or states over time.
Some memory systems, such as Mem0, overwrite or discard earlier records when new information arrives \citep{chhikara2025mem0}.
The stored memory may then lack the records needed to answer historical questions.
Other systems, such as MemOS, can retain a relevant record without marking it as current in the retrieved evidence.
In 24 of the 150 sampled failures, the evidence contained only an outdated state or left the current state unclear.
A medical memory must preserve past states and identify which state applies at the relevant time.

\textbf{(2) Subject attribution.} Longitudinal medical dialogue may contain health information about multiple subjects.
Without clear attribution, clinical information about a subject other than the patient may be mistaken for the patient's own.
In 108 of the sampled failures, the evidence contained such information.
Each clinical fact therefore needs an explicit link to the subject it describes.

\textbf{(3) Safety prioritization.} Safety information can constrain decisions across later visits.
For example, a documented allergy may affect a subsequent prescription.
In 18 of the sampled failures, we found that the evidence omitted a safety fact needed for the answer.
The memory must therefore keep safety-critical information available for later clinical decisions.

Motivated by these findings, we propose \method, a multi-layer memory with raw dialogues (L1), a clinical fact graph (L2), and a patient profile (L3).
During construction, we extract each session into typed fact nodes that record the subject and event date while preserving values and units.
Supersedes edges connect successive records of the same clinical item, including medication changes and repeated measurements.
Causal edges link related facts within a session.
At retrieval time, we rank the patient's chains as units, along with facts outside chains.
Every answer also receives the patient profile so that important details, including safety constraints, remain available to the answering model.

We evaluate \method on \bench against five existing memory systems using two backbones.
\method achieves the highest overall accuracy with both backbones, exceeding the strongest baseline by 10.5 percentage points with GPT and 10.4 with Qwen.

Our contributions are as follows:
\begin{itemize}[leftmargin=*,topsep=2pt,itemsep=1pt,parsep=0pt]
  \item We analyze failures in longitudinal medical dialogue and formulate three memory requirements: state validity, subject attribution, and safety prioritization.
  \item We introduce \method, a multi-layer memory that links subject-attributed clinical facts into dated chains. Its patient profile accompanies every answer.
  \item We show consistent accuracy gains across two backbones, with efficient memory construction and answering in the GPT setting.
\end{itemize}

\section{Related Work}
\label{sec:related}

\paragraph{Memory for LLM agents.}
External memory extends an agent's context through evidence
retrieval \citep{lewis2020rag}, memory tiers
\citep{packer2023memgpt}, fact updates
\citep{chhikara2025mem0}, and linked notes \citep{xu2025amem}.
MemOS unifies memory management \citep{li2025memos},
MIRIX coordinates specialized components \citep{wang2025mirix},
and LightMem uses compression and offline consolidation
to reduce costs \citep{fang2025lightmem}.

\paragraph{Structured and temporal memory.}
GraphRAG uses community summaries \citep{edge2024graphrag},
HippoRAG~2 retrieves passages and phrases through a graph
\citep{gutierrez2025hipporag2}, and Panini organizes memory
around entities and events \citep{rajesh2026panini}.
REMem links episodic gists with time-scoped facts
\citep{shu2026remem}; Zep/Graphiti tracks validity intervals
and invalidates conflicting relations while preserving history
\citep{rasmussen2025zep}.
\method uses type-specific clinical updates to distinguish
state changes from repeated observations.

\paragraph{Longitudinal clinical memory.}
Clinical extraction identifies negation, temporality, and
experiencer \citep{harkema2009context}, with benchmarks for
medication extraction and temporal relations
\citep{uzuner2010medication,sun2013i2b2temporal}.
LoCoMo and LongMemEval evaluate memory across sessions
\citep{maharana2024locomo,wu2024longmemeval};
\bench \benchcite focuses on longitudinal medical dialogue,
including proxy consultations.

Our work brings together persistent memory, temporal state tracking,
and clinical information extraction to maintain patient information
across visits, with explicit subject attribution and safety facts
available regardless of query relevance.
\section{Problem Setting}
\label{sec:problem}

We consider a medical assistant that follows one patient across
many dialogue sessions. After each session, it updates a memory
that it can use to answer questions about the history so far.
The dialogue may also discuss a relative's health or a general
medical topic. If the patient describes their father's medication,
that medication must remain associated with the father.

The patient's clinical state can change between visits.
A drug may be started, adjusted, and later stopped, but an earlier
dose may still be needed to answer a question about the past.
Blood pressure readings also need their dates, since a new reading
does not make an earlier measurement incorrect.
The event date may differ from the session date when a patient
reports a change that happened weeks earlier.
The assistant must use the state that applies to the time asked about.

Some information matters long after it was first reported.
An allergy mentioned early in the history may affect a later
prescription, even if the question does not mention it.
The memory must make such safety information available when
it is needed for an answer.

Let $s_t$ be the session received at step $t$, and let
$\mathcal{H}_t=(s_1,\ldots,s_t)$ be the history available
at that point. The system updates its memory as each session arrives:
\begin{equation}
    M_t=\mathcal{U}(M_{t-1},s_t).
\end{equation}
Memory updates use only sessions observed so far, without access
to future sessions or evaluation questions.
When a question $q$ arrives, the system retrieves
evidence from its current memory, then generates an answer:
\begin{equation}
    E_{t,q}=\mathcal{R}(M_t,q),
    \qquad
    \hat{a}_{t,q}=\mathcal{G}(q,E_{t,q}).
\end{equation}
Here, $\mathcal{U}$ updates memory, $\mathcal{R}$ selects and
assembles evidence, and $\mathcal{G}$ generates the answer.

We seek evidence that concerns the target patient, reflects
the time asked about, and includes the safety information
needed for the answer. We call these requirements
\textbf{subject attribution}, \textbf{state validity}, and
\textbf{safety prioritization}, respectively.
In controlled comparisons, we hold the answer model $\mathcal{G}$
fixed to examine the effects of memory construction and retrieval.

To address these requirements, we introduce \method, our memory system for tracking
patient information across visits and selecting evidence to
answer questions about the patient's history.

\section{\method}
\label{sec:method}
\begin{figure*}[t]
  \centering
  \includegraphics[width=\textwidth]{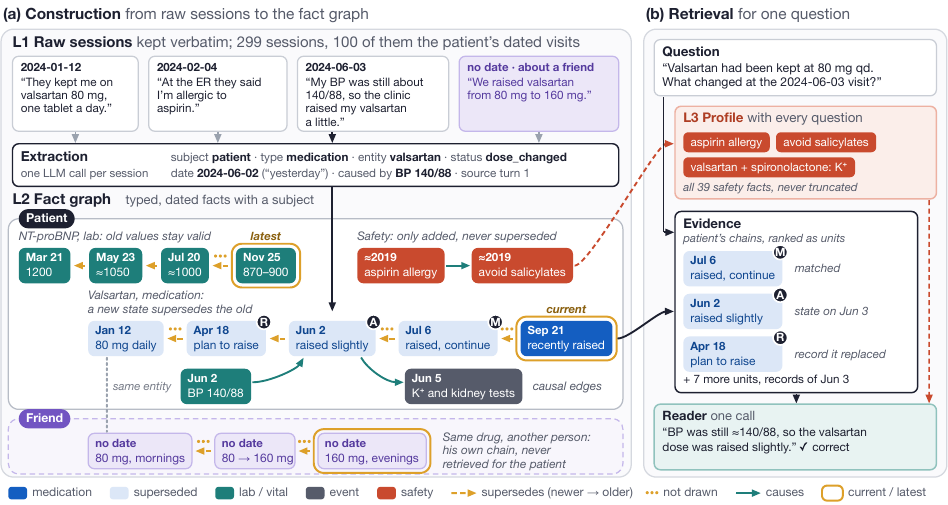}
  \caption{\method overview.
  (a)~Construction: clinical facts are extracted from each session and linked to the person they describe in the fact graph (L2). Medication, condition, lab, and vital facts form chains by entity, subject, and type, following insertion order, with the patient's and other people's facts kept separate.
  (b)~Retrieval: the system ranks the patient's records, treating each chain as one unit and each fact outside a chain as an individual unit. For selected chains, it includes the matched fact, the state at the requested time, and the preceding record. It also adds causally linked facts, records from any explicitly requested date window, and original dialogue excerpts. The answer model receives the question, the assembled evidence, and the patient profile (L3), then generates an answer in one LLM call.}
  \label{fig:overview}
\end{figure*}

\method builds a clinical fact graph from each incoming session
(Figure~\ref{fig:overview}). Raw dialogue is retained in L1 as source evidence.
L2 links extracted facts to the people they concern and the clinical items they describe, such as medications, diagnoses, or measured quantities. These items are represented as entity nodes.
At retrieval time, the patient profile (L3) is rendered from the patient's L2 records and included with every question.

Construction uses one extraction call per session. Retrieval selects graph records and adds their source text. The answer model, called the reader, receives this evidence and the profile, then answers in one LLM call.

\subsection{Memory construction (L1 to L2)}
\label{sec:method:write}

Each fact stores its type $\tau$, subject $\sigma$, clinical
entity $e$ (the item it describes), event date, and source session and turn. Its content is one sentence preserving reported numbers and units, with separate value, unit, and status fields when available. Condition facts describe diagnoses.
Medication and condition facts have status labels; safety facts have no status field. Entities and persons have separate nodes, and facts about the same entity are linked through its entity node.

\paragraph{Extraction.}
Given a session, its date, and a fixed prompt
(Appendix~\ref{app:prompt}), the extractor returns JSON with
the session's subject, clinical facts, and causal links between
those facts. The links become causal edges in L2. Each fact is assigned to the person it describes, not necessarily
the speaker. For example, a fact about the father's medication
is linked to the father's node, even when reported by the patient.
General health information stays in L1 only.
Relative dates such as ``two weeks ago'' are calculated from the session date, which is also used when the event date is unknown.
Entity names are standardized before matching to existing nodes
(Appendix~\ref{app:resolution}).

\paragraph{Chains and repeated mentions.}
Each chain groups facts with the same entity, subject, and type,
using the key $(e,\sigma,\tau)$. The patient and another person thus
have separate chains for the same drug. Medication, condition,
lab, and vital facts form chains; other types remain individual
records (Table~\ref{tab:policy}).

A new fact counts as a repeated mention if its status matches the chain tail and, after dates are removed, every Arabic numeral in its content already appears in the tail's content. It updates the existing node's last-mentioned date and mention count without creating a new node. The new wording remains in L1.

\paragraph{Chain updates.}
A fact that fails the repetition check is appended to its chain, with a supersession edge pointing to the previous tail. Chains follow insertion order, not event date, so the tail is the latest inserted record.

For medications and conditions, the previous record's validity interval ends at the new record's event date, or the session date if the event date is unavailable. Planned and active medication records share a chain, with plans labeled \texttt{planned}.

Lab and vital records are retained as historical measurements, with the latest selected by event date rather than chain position. When a change includes a previous value, such as
\emph{from 10\,mg to 20\,mg}, that value is kept in the new fact's content.

\begin{table}[t]
  \centering
  \footnotesize
  \setlength{\tabcolsep}{2.5pt}
  \begin{tabular}{@{}lll@{}}
    \toprule
    Fact type & Storage & Current record \\
    \midrule
    medication, condition & chain, insertion order & chain tail \\
    \midrule
    lab, vital            & chain, all kept        & latest by date \\
    \midrule
    symptom, event        & individual records     & -- \\
    \midrule
    safety                & append-only            & all in profile \\
    \midrule
    preference            & individual records     & -- \\
    \bottomrule
  \end{tabular}
  \caption{Storage and update policy by fact type.}
  \label{tab:policy}
\end{table}

\paragraph{Indexing.}
We index each fact's entity name, content, and available status
using BM25 \citep{robertson2009bm25} and a dense index,
retaining its subject label. L1 turn pairs are also indexed.

\subsection{Patient profile (L3)}
\label{sec:method:profile}

The profile is rendered from the patient's L2 records at retrieval
time. It includes every safety fact without truncation or
retraction: allergies, contraindications, drug interactions,
and surgical or bleeding history.

Diagnosis- and medication-chain tails and key preferences
share a fixed budget. Stopped medications are listed separately
from other medication tails, including plans.
The profile begins with the latest available session date
as the memory cutoff and is included with every question
without a relevance check.

\subsection{Retrieval}
\label{sec:method:read}

Every question uses the same retrieval pipeline.
The reader receives the full question.
Section~\ref{sec:analysis:rw} compares this pipeline with simpler retrieval controls.

\paragraph{Subject filtering and ranking.}
We rank only the patient's facts using BM25 and dense similarity.
Other people's facts remain stored under their own person nodes.
Each chain counts as one retrieval unit, while facts outside
chains count individually. Reciprocal rank fusion
\citep{cormack2009rrf} combines the rankings, and the top eight
units are retained.

\paragraph{Chain context.}
Each selected chain supplies up to three facts:
the fact that triggered retrieval (\emph{Matched}),
the record that applies to the time asked about (\emph{Anchor}),
and the record the anchor supersedes (\emph{Relation}).
For medication and diagnosis chains, the anchor's stored validity
interval must cover the explicit question date; if no date is
given, the chain tail is used. Lab and vital anchors are the
latest measurements by event date up to the requested time.

\paragraph{Expansion.}
We add up to four causal neighbors of the matched facts.
An explicit date containing a year activates the \emph{date channel},
which adds patient records from that date window, along with
medication and diagnosis records whose stored validity intervals
cover the window's start.

\paragraph{Rendering and answering.}
Each fact appears on one line with its date and available status.
We mark superseded records and chain tails.
For labs and vitals, the latest-measurement label is assigned
using event dates.
We include verbatim source windows for the top three matched facts.
The profile and retrieved evidence fill the memory slot of the
fixed answer prompt, and the reader answers in one LLM call.
Appendix~\ref{app:settings} gives the exact budgets and retrieval limits.
\section{Experiments}
\label{sec:exp}

\subsection{Setup}
\label{sec:exp:setup}

\paragraph{Benchmark.}
We evaluate \method on 967 questions from \bench \benchcite. Each history mixes about 100 patient sessions with 200 distractor sessions about general health or other people's care.
Sessions arrive in order; questions follow every ten patient sessions and concern the history available so far.
The six types are entity exact match (EEM), multiple choice (MQ), temporal location accuracy (TLA), state update accuracy (SUA), inference generation (IG), and multi-hop clinical deduction (MCD). We report per-type accuracy and their unweighted average (Avg.).

\paragraph{Baselines.}
We use five systems from \bench with their default implementations and settings. \emph{Embedding RAG} retrieves the five most similar dialogue chunks, each up to 4{,}096 tokens. \emph{Mem0} \citep{chhikara2025mem0} extracts short memories
and adds, updates, or deletes them;
\emph{MemOS} \citep{li2025memos} stores extracted memories in a vector index; and \emph{LightMem} \citep{fang2025lightmem}
compresses topic segments into time-stamped entries. \emph{REMem} \citep{shu2026remem} stores time-stamped gists and fact triples in a graph searched by an agent over several steps.

In the GPT setting, \texttt{gpt-6-luna} builds LLM-based memories and \texttt{gpt-5.1} answers questions. In the Qwen setting, Qwen3-30B-A3B-Instruct-2507 \citep{qwen2025qwen3} does both.
Embedding RAG uses no construction LLM.
All methods share the answer prompt, judge, and embedder (Appendix~\ref{app:baselines}).

\begin{table}[t]
  \centering
  \small
  \setlength{\tabcolsep}{1.7pt}
  \definecolor{firstblue}{rgb}{0.627,0.753,0.933}
  \definecolor{secondblue}{rgb}{0.859,0.906,0.976}
  \newcommand{\topa}[1]{\cellcolor{firstblue}\textbf{#1}}
  \newcommand{\topb}[1]{\cellcolor{secondblue}#1}
  \begin{tabular}{@{}lccccccc@{}}
    \toprule
    & Avg. & EEM & MQ & TLA & SUA & IG & MCD \\
    \midrule
    \multicolumn{8}{@{}l}{\textit{GPT backbone}} \\
    Emb.~RAG      & \topb{41.4} & \topb{50.5} & \topb{69.8} & 46.3 & 42.0 & \topb{22.2} & \topa{17.2} \\
    Mem0          & 40.3 & 47.5 & 65.3 & 36.2 & \topa{63.0} & 20.2 & 9.7 \\
    MemOS         & 39.1 & 46.0 & 61.8 & \topb{50.3} & \topb{61.0} & 13.6 & 2.2 \\
    LightMem      & 37.1 & 40.0 & 68.3 & 44.6 & 53.0 & 13.6 & 3.2 \\
    REMem         & 23.0 & 27.5 & 53.8 & 23.2 & 28.0 & 3.5 & 2.2 \\
    \method       & \topa{51.9} & \topa{56.5} & \topa{72.4} & \topa{68.4} & \topb{61.0} & \topa{37.9} & \topb{15.1} \\
    \quad BM25 only & 49.8 & 54.0 & 73.9 & 65.0 & 63.0 & 33.3 & 9.7 \\
    \quad built facts, BM25 & 43.6 & 51.5 & 69.3 & 54.2 & 61.0 & 19.2 & 6.5 \\
    \quad raw text, BM25 & 33.0 & 35.0 & 69.8 & 27.7 & 37.0 & 15.7 & 12.9 \\
    \midrule
    \multicolumn{8}{@{}l}{\textit{Qwen backbone}} \\
    Emb.~RAG      & \topb{35.7} & 43.0 & 48.7 & \topb{41.2} & \topb{51.0} & \topb{18.2} & \topa{11.8} \\
    Mem0          & 26.4 & 35.0 & 47.2 & 18.6 & \topb{51.0} & 4.5 & 2.2 \\
    MemOS         & 33.3 & \topb{46.0} & \topa{55.3} & 37.3 & 47.0 & 10.1 & 4.3 \\
    LightMem      & 17.6 & 22.0 & 42.7 & 14.7 & 24.0 & 1.0 & 1.1 \\
    REMem         & 16.3 & 14.5 & 45.7 & 18.6 & 19.0 & 0.0 & 0.0 \\
    \method       & \topa{46.1} & \topa{58.5} & \topb{52.8} & \topa{68.9} & \topa{69.0} & \topa{19.7} & \topb{7.5} \\
    \quad BM25 only & 46.2 & 58.5 & 53.3 & 69.5 & 70.0 & 17.2 & 8.6 \\
    \quad built facts, BM25 & 40.5 & 59.0 & 51.8 & 54.8 & 62.0 & 9.1 & 6.5 \\
    \quad raw text, BM25 & 24.5 & 24.0 & 46.2 & 24.9 & 35.0 & 10.6 & 6.5 \\
    \bottomrule
  \end{tabular}
  \caption{Accuracy (\%) with GPT
  (\texttt{gpt-6-luna} builds memory, \texttt{gpt-5.1} answers)
  and Qwen (Qwen3-30B-A3B-Instruct-2507 does both);
  \texttt{claude-sonnet-4-5} judges.
  Avg.\ is the mean across six types.
  Dark/light blue: best/second best among six methods;
  best scores are also bold.
  Indented BM25 controls (Section~\ref{sec:analysis:rw}):
  \emph{BM25 only} changes only \method's ranking;
  \emph{built facts, BM25} retrieves extracted facts without
  chain retrieval, the profile, or source windows;
  \emph{raw text, BM25} builds no memory and retrieves whole
  sessions with the benchmark's BM25 implementation.}
  \label{tab:main}
\end{table}

\subsection{Results}
\label{sec:exp:main}

\paragraph{\method leads overall, especially on TLA and IG.}
\method has the highest Avg.\ and leads on most question types,
with the largest gains over the best baseline on TLA and IG
(Table~\ref{tab:main}).
TLA requires evidence from specific times; IG depends on the
patient's conditions and medications.
These gains come from dated facts, the date channel, and the profile
(Section~\ref{sec:analysis:rw}).

\paragraph{\method also leads with an open model.}
With Qwen, \method again has the highest Avg.\
(Table~\ref{tab:main}).
All methods lose accuracy; \method's drop is similar to
Embedding RAG's, which builds memory without an LLM,
and much smaller than Mem0's and LightMem's.

\begin{figure}[t]
  \centering
  \includegraphics[width=\columnwidth]{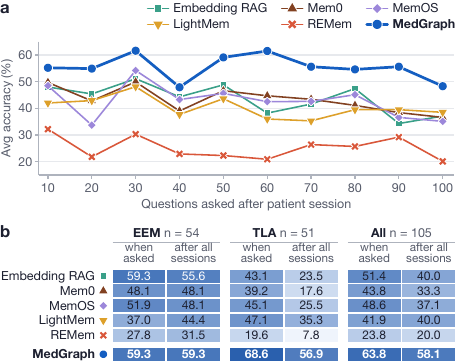}
  \caption{Accuracy as histories grow (GPT).
  (a)~Avg.\ after every ten patient sessions, using new questions.
  (b)~The same 105 early EEM and TLA questions at their original
  checkpoints and after the full history, with unchanged reference
  answers. LightMem uses one rebuilt memory for both measurements
  (Appendix~\ref{app:analysis}).}
  \label{fig:history}
\end{figure}

\paragraph{\method leads at every checkpoint.}
\method has the highest Avg.\ at all ten checkpoints
(Figure~\ref{fig:history}a).
To separate history growth from changes in questions, we repeat
the same early questions after the full history
(Figure~\ref{fig:history}b).
Embedding RAG, Mem0, and MemOS lose accuracy significantly;
\method's drop is not significant, and it remains the most
accurate (Appendix~\ref{app:analysis}).
\section{Analysis}
\label{sec:analysis}

\begin{table}[t]
  \centering
  \small
  \setlength{\tabcolsep}{1.6pt}
  \begin{tabular}{@{}lcccccc@{}}
    \toprule
    & & State & \multicolumn{2}{c}{Subject} & \multicolumn{2}{c}{Safety} \\
    \cmidrule(lr){3-3}\cmidrule(lr){4-5}\cmidrule(l){6-7}
    & Avg. & Acc. & \shortstack{Polluted\\entries} & \shortstack{Polluted\\questions} & Missing & Acc. \\
    \midrule
    Emb.~RAG       & 41.4 & 49.0 & 12.2 & 22.8 & 38.2 & 44.0 \\
    Mem0           & 40.3 & 55.4 & 22.4 & 55.6 & 51.7 & 44.9 \\
    MemOS          & 39.1 & 53.4 & 17.1 & 40.4 & 59.6 & 39.7 \\
    LightMem       & 37.1 & 52.9 & 21.8 & 74.1 & 74.2 & 40.8 \\
    REMem          & 23.0 & 35.8 & 27.3 & 63.6 & 80.9 & 25.6 \\
    \method        & \textbf{51.9} & \textbf{60.3} & \textbf{0.0} & \textbf{0.1} & \textbf{1.1} & \textbf{60.4} \\
    \midrule
    $-$ subject filter & 49.6 & 57.8 & 15.9 & 57.8 & 1.1 & 57.1 \\
    $-$ profile & 49.3 & 56.4 & 0.0 & 0.1 & 49.4 & 56.0 \\
    \bottomrule
  \end{tabular}
\caption{The three challenges (\%, GPT). State: accuracy on 204 changed-state questions (Appendix~\ref{app:baselines}). Subject: percentages of evidence entries about other people and questions with such evidence (lower is better; Appendix~\ref{app:baselines}). Safety: the missing-fact rate on 89 questions with identified required safety facts, and mean MQ and IG accuracy on 180 safety-dependent questions. Avg.\ follows Table~\ref{tab:main}. Bold: best among six systems. Bottom rows omit one component.}
  \label{tab:challenges}
\end{table}

\begin{table*}[t]
  \centering
  \small
  \renewcommand{\arraystretch}{1.15}
  \setlength{\tabcolsep}{5pt}
  \begin{tabularx}{\textwidth}{@{}l>{\raggedright\arraybackslash}X>{\raggedright\arraybackslash}p{0.24\textwidth}@{}}
    \toprule
    \multicolumn{3}{@{}p{\textwidth}@{}}{\textbf{Question} (patient 4, IG): ``Doctor, my knee pain is back. Can I take a painkiller again?''} \\
    \multicolumn{3}{@{}p{\textwidth}@{}}{\textbf{Gold}: Avoid ibuprofen, which the doctor stopped; with perindopril and spironolactone it risks kidney damage and high potassium.} \\
    \midrule
    & \textbf{Evidence} & \textbf{Answer} \\
    \midrule
    LightMem \xmark & \evbad{``Took ibuprofen 200\,mg as needed for about a year''} (Jan 3) and ``the doctor told the patient to stop ibuprofen'' (Jan 5), with no mark of which is current; none names perindopril or spironolactone. & \evbad{``If the pain is very bad, a one-off 200\,mg of ibuprofen is acceptable.''} \\
    MemOS \xmark & Knee-pain and exercise memories. Its memory holds the order to stop ibuprofen and the interaction warning, but neither is retrieved. & No drug record; an old painkiller ``can be considered''. \\
    REMem \xmark & General advice and another person's records, e.g., \evbad{``continue mecobalamin to support nerve recovery''}. Its stored interaction warning is not retrieved. & Asks whether the user is ``Xiao Zhou, who used celecoxib''. \\
    \midrule
    \method\ \cmark & Current medications: perindopril, spironolactone, and valsartan, all \evgood{``in use''} with dates. Safety: \evgood{``[interaction] ibuprofen: with perindopril and spironolactone, adds kidney and potassium risk; do not keep taking it''} (2024-01-03). & \evgood{``With perindopril, valsartan, and spironolactone, do not take ibuprofen on your own.''} \\
    \bottomrule
  \end{tabularx}
  \caption{A question that needs the current state and a safety fact (translated and shortened). Red: misleading; blue: what the correct answer needs. More examples are in Appendix~\ref{app:examples}.}
  \label{tab:case}
\end{table*}

\subsection{Memory construction and retrieval}
\label{sec:analysis:rw}

\begin{figure}[t]
  \centering
  \includegraphics[width=\columnwidth]{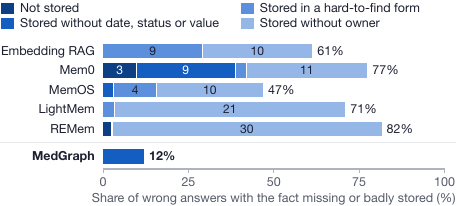}
  \caption{Storage and indexing issues in sampled failures. Percentages use 31, 31, 32, 31, 38, and 25 retained cases in the plotted order. Numbers inside bars are counts. LightMem uses a rebuilt memory  (Appendix~\ref{app:analysis}).}
  \label{fig:stage}
\end{figure}

\paragraph{Most sampled baseline failures come from missing or badly stored facts.}
We check sampled wrong answers of each system against its stored memory and retrieved evidence (Appendix~\ref{app:analysis}).
In about two thirds of the baseline cases, the needed fact is not stored, stored incompletely, stored in a hard-to-find form, or stored with no owner (Figure~\ref{fig:stage}).
In most of \method's sampled failures, the needed fact is stored intact.

\paragraph{Fact construction accounts for more than half of the gain over raw text.}
To separate construction from retrieval, we rank raw sessions and \method's extracted facts with BM25 alone (Table~\ref{tab:main}).
With both backbones, these facts alone give more than half of the full method's gain over raw sessions, and they score higher than raw sessions for every patient.
Each fact is a short statement about one person and carries its date, so BM25 can match the record a question needs, while a raw session is long and covers many topics.
Replacing \method's fused BM25 and dense ranking with BM25 alone, while keeping the rest of the retrieval, changes accuracy little, so the scoring function accounts for little of the lead.

\paragraph{Retrieval helps most on TLA and IG.}
Compared with the facts ranked by BM25, the full retrieval (chain retrieval, the date channel, the profile, and source windows) gains most on IG and TLA with GPT (Table~\ref{tab:main}).
TLA questions ask when an event was recorded or what was recorded on a given date.
When a question names a date, the date channel adds the records of that day, and each chain is read at that date.
IG questions depend on the patient's own diagnoses, drugs, and allergies, which the profile gives with every question (Table~\ref{tab:case}).

\subsection{The three challenges}
\label{sec:analysis:challenges}

\paragraph{State validity: \method is the most accurate on changed-state questions.}
\method answers questions about changed states more accurately than every baseline (Table~\ref{tab:challenges}); these are the 100 SUA questions and the 104 MQ and IG questions whose benchmark trap is a changed state or dose (Appendix~\ref{app:baselines}).
Most of this gain comes from construction: ranked by BM25 alone, the extracted facts answer these questions significantly more often than raw sessions (Appendix~\ref{app:baselines}).
Each record has a status and a date, so the reader can tell which state is current. In Table~\ref{tab:case}, LightMem retrieves the earlier ibuprofen use and the order to stop it as two unmarked entries.

\begin{figure}[t]
  \centering
  \includegraphics[width=\columnwidth]{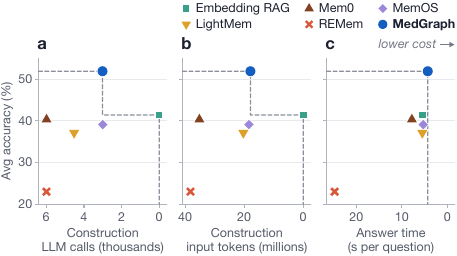}
  \caption{Accuracy against cost (GPT backbone): (a, b)~LLM calls and input tokens spent constructing memory; (c)~seconds per question, the wall-clock time from the question to the answer string, including the method's retrieval and the reader's generation and excluding judging. The top right is best; dashed lines mark the Pareto frontier.}
  \label{fig:cost}
\end{figure}

\paragraph{Subject attribution: \method keeps other people's facts out of the evidence.}
The baselines often retrieve other people's health facts, while \method almost never does (Table~\ref{tab:challenges}).
Embedding RAG returns chunks of consultations about other people, and the textual memories have no subject label to filter on.
During construction, \method assigns all but 9 of the 11{,}039 facts from consultations about other people to someone other than the patient, and the subject filter keeps them out of the evidence.
Without the filter, other people's facts appear in the evidence of more than half of the questions, and accuracy drops.
In Table~\ref{tab:case}, REMem retrieves another person's records.

\paragraph{Safety prioritization: the profile supplies the safety facts.}
On questions whose answer depends on an allergy, a contraindication, or a drug interaction, the baselines often miss the needed safety fact, while \method's evidence almost always contains it (Table~\ref{tab:challenges}).
The profile lists every safety fact with each question, so the reader sees it even when the question does not mention it.
Without the profile, the fact is often missing, and accuracy drops.
In Table~\ref{tab:case}, MemOS stores the ibuprofen warning but does not retrieve it.

\subsection{Cost and efficiency}
\label{sec:analysis:cost}

\paragraph{\method is the only LLM-built memory on the accuracy--cost frontier.}
\method makes one LLM call per session to build its memory, about as many as MemOS and half as many as Mem0 and REMem (Figure~\ref{fig:cost}).
It also has the shortest mean answering time: one retrieval pass and one LLM call per question, while REMem first searches with several LLM calls.

\section{Conclusion}
\label{sec:conclusion}

We presented \method, a graph memory that tracks clinical state changes, attributes facts to their subjects, and includes safety facts with every question. On \bench, it leads with both GPT and Qwen by 10.5 and 10.4 points over the strongest baseline.
Compared with retrieval over raw sessions, more than half of the gain comes from fact construction and the rest from retrieval.
 In the GPT setting, \method builds memory with one extraction call per session, uses about half the construction input tokens of Mem0 and REMem, and has the shortest mean answering time. Future work could link causes across visits to support
multi-hop questions.

\section*{Limitations}

We evaluate \method on one Chinese benchmark of longitudinal medical dialogue; evaluating it in other languages and clinical settings is left to future work.
\method relies on an LLM to extract facts from each session, so the quality of its memory depends on the extractor.
Causal links are drawn within a session only.
Chains are ordered by when a fact is recorded, not by when the event happened, so a late report of an earlier event or a planned change becomes the chain tail.
Safety facts are never removed, so an allergy that is later ruled out, or a warning about a drug the patient has since stopped, stays in the profile.
Following the benchmark's protocol, four question types are scored by an LLM judge.

\section*{Ethics Statement}

\method is a research prototype and is not validated for clinical use.
A memory that drops a current medication, keeps a stopped one, or attributes a relative's diagnosis to the patient can lead to harmful advice; any deployment would need clinical review of what the memory shows.
All experiments use the synthetic patients and dialogues of \bench; no real patient data was used.

\section*{Use of AI Assistants}

AI assistants were used for language polishing and grammar correction.
All text, code, and results were reviewed by the authors.

\bibliography{medgraph}
\clearpage
\appendix
\lstdefinestyle{promptstyle}{basicstyle=\ttfamily\scriptsize, columns=fullflexible, keepspaces=true,
  breaklines=true, breakatwhitespace=true, breakindent=1.2em,
  postbreak=\mbox{\textcolor{gray}{$\hookrightarrow$}\space}, showstringspaces=false,
  literate={★}{{\ding{72}}}1}
\tcbset{promptframe/.style={breakable, enhanced, colback=white, colframe=black, boxrule=0.8pt, arc=2mm,
  left=2mm, right=2mm, top=3mm, bottom=1mm, coltitle=white, colbacktitle=black, fonttitle=\bfseries\small,
  attach boxed title to top left={xshift=2mm,yshift=-2mm}, boxed title style={sharp corners, size=small}}}
\newtcblisting{promptbox}[2][]{promptframe, listing only, listing options={style=promptstyle}, title=#2, #1}
\newtcolorbox{promptboxtext}[2][]{promptframe, title=#2, #1}

\section{Extraction prompt}
\label{app:prompt}

The experiments use a Chinese prompt; its English translation is given first, followed by the original.
The examples for subject attribution use placeholder names (disease~A, drug~B, metric~C).

\begin{promptbox}{Extraction prompt (English translation)}
You are a medical dialogue information extractor. Extract structured clinical facts from the doctor-patient dialogue below and output strict JSON.

[Dialogue date] {date}
[Dialogue]
{dialogue}

[Requirements]
0. [Decide first, then extract] Whose health is this dialogue about? Check whom each piece of health information in the patient's words is attached to: "my mother / my father / my son / my daughter / my spouse / my friend / she / he" + symptom / medication / test / measurement -> the owner of these health facts is THAT person, not the speaker. (!) The speaker tag "Patient" does not make the patient the subject of a fact: in a family health consultation the patient only relays the information, and the subject is family:<relation>.
1. Label the subject of each fact separately. Test: whose body has the symptom, whose mouth takes the drug, whose body the measurement belongs to.
   - "patient" = the health fact belongs to the patient
   - "family:<relation>" = the health fact belongs to a family member (e.g. "family:father")
     * "My mother has had disease A for years and keeps taking drug B" -> the disease and the drug are the mother's -> family:mother
     * "My husband's hospitalization stressed me, and my own metric C got worse" -> metric C is the patient's own body -> patient (the husband is only a trigger; no health fact of his is stated)
     * The test does not depend on who says it: the doctor says "your wife was diagnosed with disease A" -> the disease is the wife's -> family:wife
   - "none" = general health knowledge not tied to any person
   [Micro-example] Dialogue: "Patient: Doctor, my son is 15, he has disease A and is on drug B, and he says his symptoms have not improved."
   Correct: session_subject="family:son"; both facts have subject "family:son" (medication_status drug B active; symptom not improved)
   Wrong: subject "patient" <- the speaker is the patient, but the disease and the drug are the son's.
2. type (one per fact):
   - medication_status: start / stop / dose change / planned / self-medication; status required: active | stopped | dose_changed | planned
   - condition: diagnosis status; status required: suspected | confirmed | misdiagnosed | resolved | chronic | worsened | improved
   - lab_result: lab, imaging, or antibody result; keep the exact value and unit in value
   - vital: weight, blood pressure, home glucose, etc.; keep the exact value
   - symptom: onset / worsening / relief
   - safety: ★ safety-critical information, always extracted as its own fact: allergy, contraindication or interaction, prior surgery or bleeding (both surgical_history); safety_kind required: allergy | contraindication | interaction | surgical_history. Better too many than one missed.
   - preference: lifestyle, schedule, diet, cost and care preferences (cost concerns, reluctance to take leave, adherence)
   - event: other meaningful clinical or life events
3. entity: the core entity (drug / disease / test / metric name), or null. Use the generic drug name (e.g. "empagliflozin"), without dose.
4. date: YYYY-MM-DD. Resolve "two weeks ago" or "last month" against the dialogue date; for long-standing history give the year, e.g. "2019-01-01"; null if it cannot be determined (defaults to the dialogue date).
5. is_state_change: whether the fact changes a previously known state (stopping or switching a drug, a revised diagnosis, a relieved symptom). If so, replaces briefly describes the old state (e.g. "empagliflozin being taken"); otherwise null.
6. causes: indexes (from 0) of other facts in this output that caused this fact; [] if none.
7. content: one sentence that fully describes the fact, keeping the key numbers verbatim. confidence: 0-1.
8. evidence_turn: the dialogue turn the fact mainly comes from (from 1).

[Output format] Output JSON only, no other text:
{"session_subject": "patient|family:<relation>|mixed|none",
 "facts": [
  {"type": "medication_status", "subject": "patient",
   "entity": "empagliflozin",
   "content": "Stopped empagliflozin as advised; had been taking it on their own for about 8 months",
   "value": null, "unit": null, "status": "stopped",
   "date": "2024-01-15", "is_state_change": true,
   "replaces": "empagliflozin being taken", "safety_kind": null,
   "causes": [1], "evidence_turn": 3, "confidence": 0.95},
  {"type": "safety", "subject": "patient", "entity": "empagliflozin",
   "content": "Empagliflozin with insulin markedly raises the risk of euglycemic ketoacidosis; contraindicated",
   "value": null, "unit": null, "status": null,
   "date": "2024-01-15", "is_state_change": false, "replaces": null,
   "safety_kind": "contraindication",
   "causes": [], "evidence_turn": 3, "confidence": 0.95}]}
\end{promptbox}

\ifdefined\nolinenumbers\nolinenumbers\fi
\begin{promptboxtext}{Extraction prompt (Chinese original)}\begin{CJK*}{UTF8}{gbsn}\ttfamily\scriptsize\raggedright\setlength{\parindent}{0pt}\noindent
你是一个医疗对话信息抽取器。请从下面这段医患对话中抽取结构化临床事实,输出严格的 JSON。 \\
\mbox{} \\
【对话日期】\{date\} \\
【对话内容】 \\
\{dialogue\} \\
\mbox{} \\
【抽取要求】 \\
0. 【第一步,先判断再抽取】这段对话在谈论【谁的健康】? \\
\hspace*{1.5em}逐条检查患者话里的健康信息挂在谁身上:出现"我妈/我爸/我儿子/我女儿/我爱人/我朋友/她/他" \\
\hspace*{1.5em}+症状/用药/检查/指标 → 这些健康事实的主人是【那个人】,不是说话人。 \\
\hspace*{1.5em}(!) 说话人标签是"Patient"不代表事实主体是 patient: \\
\hspace*{2.5em}家属健康咨询里,患者只是【转述者】,事实主体 = family:关系。 \\
1. 每条事实单独标注 subject(主体)。判据:【症状长在谁身上、药吃进谁嘴里、指标是谁的身体】: \\
\hspace*{1.5em}- "patient" = 该健康事实属于患者本人 \\
\hspace*{1.5em}- "family:关系" = 该健康事实属于家人(如 "family:父亲") \\
\hspace*{3.5em}· "我妈妈得了A病好几年,一直在用B药" → 病和用药都是妈妈的 → family:母亲 \\
\hspace*{3.5em}· "丈夫住院让我压力大,我自己的C指标也变差了" → C指标是患者自己的身体 → patient \\
\hspace*{4.5em}(丈夫只是诱因,他本人没有健康事实被陈述) \\
\hspace*{3.5em}· 该判据与【谁说的】无关:医生说"你妻子确诊了A病" → 病是妻子的 → family:妻子 \\
\hspace*{1.5em}- "none" = 与任何具体个人无关的通用健康科普知识 \\
\hspace*{1.5em}【微例】对话: "Patient: 医生,我儿子15岁,得了A病,在用B药,他说症状还没好转。" \\
\hspace*{2.5em}正确: session\_subject="family:儿子"; 两条事实的 subject 都是 "family:儿子" \\
\hspace*{3.5em}(medication\_status B药 active; symptom 症状未好转) \\
\hspace*{2.5em}错误示范: subject 写成 "patient" ← 说话人是患者,但病和用药都是儿子的。 \\
2. type 取值(每条选一个): \\
\hspace*{1.5em}- medication\_status: 用药(开始/停用/调量/计划用/自行服用),status 必填: \\
\hspace*{2.5em}active|\allowbreak{}stopped|\allowbreak{}dose\_changed|\allowbreak{}planned \\
\hspace*{1.5em}- condition: 疾病诊断状态,status 必填: suspected|\allowbreak{}confirmed|\allowbreak{}misdiagnosed|\allowbreak{}resolved|\allowbreak{}chronic|\allowbreak{}worsened|\allowbreak{}improved \\
\hspace*{1.5em}- lab\_result: 检查结果(化验/影像/抗体),value 尽量保留原文精确数值和单位 \\
\hspace*{1.5em}- vital: 生理指标(体重/血压/自测血糖等),value 保留精确数值 \\
\hspace*{1.5em}- symptom: 症状(出现/加重/缓解) \\
\hspace*{1.5em}- safety: ★安全关键信息,必须单独抽出:过敏史(allergy)、用药禁忌/相互作用 \\
\hspace*{2.5em}(contraindication/interaction)、既往手术或出血史(surgical\_history), \\
\hspace*{2.5em}safety\_kind 必填。宁多勿漏! \\
\hspace*{1.5em}- preference: 生活方式/作息/饮食/经济与就医偏好(费用顾虑/怕请假/依从性) \\
\hspace*{1.5em}- event: 其他有意义的临床或生活事件 \\
3. entity: 该事实的核心实体名(药名/疾病名/检查名/指标名),没有则 null。 \\
\hspace*{1.5em}药名用规范通用名(如"恩格列净"),不要带剂量。 \\
4. date: 事实发生日期 YYYY-MM-DD;对话里说"两周前/上个月"要以对话日期为锚换算; \\
\hspace*{1.5em}长期既往史给年份如 "2019-01-01";实在无法确定给 null(默认=对话日期)。 \\
5. is\_state\_change: 该事实是否改变了此前已知状态(如停药/换药/诊断修订/症状缓解)。 \\
\hspace*{1.5em}若是,replaces 写被取代的旧状态简述(如"恩格列净 正在服用"),否则 null。 \\
6. causes: 本次输出列表中,导致该事实的其他事实的下标数组(从0计,没有给 [])。 \\
7. content: 一句话完整描述该事实,保留关键数值原文。confidence: 0\textasciitilde{}1。 \\
8. evidence\_turn: 该事实主要出自对话的第几轮(从1计)。 \\
\mbox{} \\
【输出格式】只输出 JSON,不要任何其他文字: \\
\{"session\_subject": "patient|\allowbreak{}family:关系|\allowbreak{}mixed|\allowbreak{}none", \\
\hspace*{1.0em}"facts": [ \\
\hspace*{2.0em}\{"type": "medication\_status", "subject": "patient", "entity": "恩格列净", \\
\hspace*{3.0em}"content": "按医嘱停用恩格列净,此前自行连续服用约8个月", \\
\hspace*{3.0em}"value": null, "unit": null, "status": "stopped", \\
\hspace*{3.0em}"date": "2024-01-15", "is\_state\_change": true, \\
\hspace*{3.0em}"replaces": "恩格列净 正在服用", "safety\_kind": null, \\
\hspace*{3.0em}"causes": [1], "evidence\_turn": 3, "confidence": 0.95\}, \\
\hspace*{2.0em}\{"type": "safety", "subject": "patient", "entity": "恩格列净", \\
\hspace*{3.0em}"content": "恩格列净与胰岛素联用会显著增加正常血糖酮症酸中毒风险,医嘱禁用", \\
\hspace*{3.0em}"value": null, "unit": null, "status": null, \\
\hspace*{3.0em}"date": "2024-01-15", "is\_state\_change": false, "replaces": null, \\
\hspace*{3.0em}"safety\_kind": "contraindication", "causes": [], "evidence\_turn": 3, \\
\hspace*{3.0em}"confidence": 0.95\} \\
\hspace*{1.0em}]\}
\end{CJK*}\end{promptboxtext}
\ifdefined\linenumbers\linenumbers\fi

\section{Entity resolution details}
\label{app:resolution}
The surface name is first canonicalized through a seed dictionary of generic drug names and drug classes, then matched against existing entities of the same type: an exact match on the normalized name or a stored alias reuses the node; otherwise the node with the highest character-bigram overlap is reused only if the overlap is at least $0.75$; otherwise a new entity is created.
Class-level entries in the seed dictionary (a drug class listed with its members) merge members into the class.

\section{Retrieval settings}
\label{app:settings}
Each ranked list (dense and BM25) is scanned until it yields 32 distinct units or 256 facts, and the top 8 units after reciprocal rank fusion are kept.
Causal expansion adds at most 4 facts.
Date-anchored retrieval adds at most 12 records in the window and at most 4 states in force at its start.
The profile budget is about 1{,}600 characters, of which the safety part is never trimmed.
Source windows are shown for the top 3 matched facts, and if fewer than 2 facts are retrieved, the top 4 raw turn pairs are added.
The same settings are used for every question and every patient.

\section{Baselines and evaluation details}
\label{app:baselines}

\paragraph{Protocol.}
We use the first ten patients of \bench by ID and all 967 of their questions in the Chinese mixed setting.
All methods run in the benchmark's harness with the same reader (\texttt{gpt-5.1}, temperature 0.3), the same answer prompt, the same judge (\texttt{claude-sonnet-4-5}), and the same local embedder (bge-small-zh-v1.5; \citealp{xiao2023bge}).
Every LLM call that builds memory, by our extractor and by the baselines' memory builders, uses \texttt{gpt-6-luna}.
Memory is built online as sessions arrive, and questions are answered from memory alone.
EEM is scored by string containment, MQ by option match, and the other four types by an LLM judge with the benchmark's prompts.

\paragraph{Baseline settings.}
All baselines are the implementations shipped with \bench and run through its harness, with their default configurations (Table~\ref{tab:baselines}).
Apart from the REMem rebuild described below, we change only the model that builds memory, so that every method builds memory with the same model.
Retrieval budgets are the defaults of each method and are not equalized.
REMem's prompts are in English, so it writes English paraphrases of the Chinese sessions, and its retrieval agent runs on the reader model.
As released, the REMem adapter rebuilds REMem at every checkpoint from only the sessions since the previous checkpoint, so REMem would remember only the latest ten visits; we rebuild it from all sessions so far, reusing earlier extraction responses so that each session is still extracted once.
In the Qwen setting (Table~\ref{tab:main}), Qwen3-30B-A3B-Instruct-2507 (FP8 weights, served with vLLM) replaces both \texttt{gpt-6-luna} as the memory builder and \texttt{gpt-5.1} as the reader, at the same reading temperature of 0.3; everything else is as in Table~\ref{tab:baselines}.

\begin{table}[h]
  \centering
  \small
  \setlength{\tabcolsep}{2.5pt}
  \begin{tabular}{@{}lll@{}}
    \toprule
    Method & Builder & Evidence per question \\
    \midrule
    Embedding RAG & none & top 5 chunks ($\le$4{,}096 tok.) \\
    Mem0          & \texttt{gpt-6-luna} & top 5 memories \\
    MemOS         & \texttt{gpt-6-luna} & top 5 memories \\
    LightMem      & \texttt{gpt-6-luna} & top 20 memories \\
    REMem         & \texttt{gpt-6-luna} & top 5, gist agent \\
    \method       & \texttt{gpt-6-luna} & 8 units, $\le$4 causal, dates \\
    \bottomrule
  \end{tabular}
  \caption{Baseline settings. The reader (\texttt{gpt-5.1}), answer prompt, judge (\texttt{claude-sonnet-4-5}), and embedder (bge-small-zh-v1.5) are shared by all methods.}
  \label{tab:baselines}
\end{table}

\paragraph{Judging.}
TLA, SUA, and IG are judged with the benchmark's prompts, and MCD with its node-level prompt.
A judge reply that cannot be parsed counts as wrong, as in the benchmark; this affects 11 of the 568 judged answers of \method and 11 to 25 for the baselines.

\paragraph{Statistics.}
Per-question tests are exact McNemar tests, and differences are computed before rounding.
In the main setting, \method is 10.5 points above embedding retrieval, the strongest baseline; against every baseline, it is higher on at least 9 of the 10 patients, and $p<10^{-9}$.
With the open model, it is 10.4 above embedding retrieval (9 of 10 patients) and more than 12 points above the other baselines, with $p<10^{-9}$ for every baseline.
Removing the subject filter or the profile lowers Avg.\ significantly ($p=0.02$ and $p=0.008$; Table~\ref{tab:challenges}).
With GPT, \method's extracted facts ranked by BM25 alone are 10.6 points above the raw sessions ranked by BM25 (10 of 10 patients; $p<10^{-9}$); with Qwen, they are 16.0 above (10 of 10 patients).
On the 204 changed-state questions with GPT, the same facts answer 55.4\% correctly, against 42.6\% for the raw sessions ($p<0.001$).

\paragraph{Tracing relatives' facts (Table~\ref{tab:challenges}).}
Each benchmark session is the patient's own consultation, a consultation about a relative or friend, or a general health question, and its header says which.
We trace every retrieved item to the session it was written from.
Embedding retrieval returns session text, which includes the header, and \method's facts and snippets record their session.
Mem0 and MemOS log the items written from each session; a Mem0 item counts as a relative's when any session that wrote or later updated it is a relative's consultation.
REMem links each stored gist to its source passage.
LightMem stamps items from the undated sessions, which are exactly the non-patient ones, with the time of the run; we assign each such item to the most similar non-patient session by BM25.
An item from a relative's consultation counts only if it states a health fact of a particular person, not just general advice or the user's worries; \texttt{gpt-6-luna} decides this for every such item.

\paragraph{Changed-state questions (Table~\ref{tab:challenges}).}
The 204 questions are all 100 SUA questions and the 104 MQ and IG questions whose benchmark trap type concerns a state or dose that has changed: \texttt{temporal\_state} (69 MQ), \texttt{temporal\_change} (22 IG), and \texttt{dosage\_adjustment} (13 IG).
The set is defined only by the benchmark's annotations.

\paragraph{Safety-dependent questions (Table~\ref{tab:challenges}).}
The 180 questions are the MQ and IG questions whose benchmark trap type is a contraindication, an allergy, or a drug interaction.
In 97 of them the annotation names the safety item; we keep the 89 whose item is a single drug, drug class, or allergen to avoid, and drop eight whose item is not a drug or allergen.
A safety fact counts as present if the drug or allergen, or a synonym from a fixed list, appears within 40 characters of a safety word (allergy, contraindication, avoid, interaction, and similar) in the reader's context.
For embedding retrieval we restore the full retrieved sessions, since its logs keep only the first 500 characters of each.
REMem writes English memories, and its logs keep only the final gists, not the triples and earlier gists its agent also reads.
For REMem we therefore rerun its retrieval agent on the memory stored at each question's checkpoint, with the same models and settings, record every gist and triple it reads, and check each item for the safety fact (the drug or allergen, including English names and class members, stated with a safety relation).

\paragraph{Ablation runs.}
Each \method variant rebuilds the memory from the same cached extraction outputs, in both model settings, so only the retrieval side differs.
For the subject-filter ablation, a question whose rendered evidence is identical to that of the full model keeps the full model's answer and judgment; the other 771 questions are answered and judged again.
\emph{BM25 only} keeps the full retrieval but ranks by BM25 alone.
\emph{Built facts, BM25} keeps the same memory and the subject filter but shows only the top 18 facts ranked by BM25 (18 is the mean number of fact lines, 17.8, that the full retrieval shows per question), with no chain units or packs, date channel, causal expansion, profile, source windows, memory-cutoff line, or raw-text fallback.
\emph{Raw text, BM25} is the benchmark's BM25 retriever over whole raw sessions (jieba tokens, top 5, up to 10k tokens) and builds no memory.

\section{Additional analysis}
\label{app:analysis}

\paragraph{Failure analysis (Figure~\ref{fig:problem}a).}
We sample 30 wrong answers from each baseline (150 in total).
For each answer, we label its main cause (e.g., another person's fact, an outdated state, a missing safety fact, a missing record) and add two marks: whether the evidence holds a health fact of someone other than the patient, and whether it holds two values or states of the same item of the patient with no date or wording that shows which is current.
A failure involves state validity if it has the second mark or its main cause is an outdated state; subject attribution if it has the first mark or its main cause is another person's fact; and safety prioritization if its main cause is a missing safety fact, or if it is one of the 89 questions with identified required safety facts (Appendix~\ref{app:baselines}) and its evidence lacks the needed fact.
Of the 150, 24 involve state validity, 108 subject attribution, and 18 safety prioritization; 129 involve at least one.

\paragraph{Storage check (Figure~\ref{fig:stage}).}
We sample 40 wrong answers from each baseline and from \method with a fixed seed (240 in total) and check what each method stored for the needed fact, as of the question's checkpoint, in this order: the gold answer or the judge is wrong; the fact is not stored; it is stored without its date, its current-state mark, or its exact value; it is stored in a hard-to-find form; it is stored with no owner, that is, it is missing from the evidence while the evidence holds other people's facts that the memory keeps with no owner; or it is stored intact.
A form counts as hard to find only with concrete evidence: the fact is inside a long entry that covers several other topics, or the stored wording lacks the key terms that the visit and the question share.
All labels are assigned in one pass per method with the same written criteria.
Figure~\ref{fig:stage} leaves out the answers whose gold answer or judge is wrong (9, 9, 8, 9, and 2 for the baselines in table order, and 15 for \method), since they are not failures of the memory; \method has the fewest wrong answers, so such cases make up a larger share of its sample.
The memory of the LightMem run was not kept, so its answers are checked against the rebuilt memory described below, built from the same sessions with the same model.

\paragraph{Fixed early questions.}
For Figure~\ref{fig:history}, we take the 105 EEM and TLA questions asked after sessions 10--30 whose answer does not change over time, leaving out questions about the current or recent state.
We answer them again with the same reader, prompt, and judge from each method's memory after all sessions: \method and embedding retrieval rebuild it from the same cached inputs, and Mem0, MemOS, and REMem load the memory they stored in the main run.
LightMem's store was not kept, so we rebuild it once with the same models and settings and answer each question twice from this rebuild, when its checkpoint is reached and after all sessions.
Accuracy on these questions changes by $-11.4$ points for embedding retrieval (exact McNemar $p=0.017$), $-10.5$ for Mem0 ($p=0.043$), $-11.5$ for MemOS ($p=0.012$), $-1.9$ for LightMem ($p=0.82$; $-11.8$ on TLA and $+7.4$ on EEM), $-3.8$ for REMem ($p=0.39$), and $-5.7$ for \method ($p=0.15$).

\paragraph{Where \method fails.}
Of the 39 wrong state-update answers, only one comes from the state representation: a later session restated an old lab value, the extractor dated it as a new measurement, and the chain treated that record as current.
Eight are not memory errors (four correct answers were misjudged, and four are disagreements with the reference answer), and the remaining 30 are retrieval or reading errors.

\section{Examples}
\label{app:examples}

Tables~\ref{tab:exA}--\ref{tab:exC} show three questions from the runs, one per property.
Each table puts the evidence and answer of the baselines that fail next to those of \method.
Evidence and answers are translated from Chinese and shortened; omitted text is marked with \dots\ and the facts quoted for \method are lines of its rendered evidence.
Red marks evidence that misleads the reader; blue marks the evidence the correct answer needs.

\newcolumntype{Y}{>{\raggedright\arraybackslash}X}
\newcommand{\exhead}[1]{\rowcolor{gray!12}\multicolumn{2}{@{\hspace{5pt}}>{\raggedright\arraybackslash}p{\dimexpr\textwidth-10pt}@{\hspace{5pt}}}{#1}}
\newcommand{\exwide}[1]{\multicolumn{2}{@{\hspace{5pt}}>{\raggedright\arraybackslash}p{\dimexpr\textwidth-10pt}@{\hspace{5pt}}}{#1}}
\newenvironment{extable}{\small\setlength{\aboverulesep}{0pt}\setlength{\belowrulesep}{0pt}\renewcommand{\arraystretch}{1.3}%
  \tabularx{\textwidth}{@{\hspace{5pt}}Y!{\color{gray!45}\vrule}Y@{\hspace{5pt}}}}{\endtabularx}

\begin{table*}[p]
\centering

\begin{extable}
  \toprule
  \exhead{\textbf{Question} (patient 9, inference): ``Doctor, I am always tired in the afternoon. Is it the medicine? Should I stop it for now?''} \\
  \exhead{\textbf{Gold}: do not stop; the afternoon fatigue was there before the diabetes drugs were started, so it is unlikely to be their side effect.} \\
  \midrule
  \textbf{Embedding RAG, MemOS} \xmark & \textbf{\method} \cmark \\
  \emph{Evidence.} Embedding RAG retrieves three noise sessions---two general health consultations and \evbad{one about the father's blood-pressure drugs}---and none of the patient's own. Three of MemOS's five memories are about the father's blood pressure, e.g., \evbad{``if the father's systolic pressure falls to about 110\,mmHg with dizziness, indapamide can be paused for a day''}.
  &
  \emph{Evidence.} Profile: \evgood{``DPP-4 inhibitor: active, started 2024-01-10''}; ``metformin: active, 1{,}500\,mg/day on a new regimen for about two weeks'' (both as of 2024-01-10). Matched facts: \evgood{``(2024-01-04) afternoon fatigue, like being drained, worst at 3--4\,pm''}; ``(2024-01-08) afternoon fatigue and poor concentration''; ``(2024-01-10) afternoon fatigue still present''. No fact about the father appears. \\
  \emph{Answer} (both discuss the father's drugs). ``\dots not likely caused by \evbad{your father's current drugs (valsartan/amlodipine)} \dots'' (Embedding RAG).
  &
  \emph{Answer.} The fatigue was reported before the drug change, so it is unlikely to be a side effect; do not stop on your own. \\
  \midrule
  \exwide{\textbf{Why.} The subject filter keeps the father's sessions out and the profile gives the patient's own diabetes drugs, so the answer is about the right person's drugs. The dated fatigue records in the evidence start on 2024-01-04, before the DPP-4 inhibitor is recorded as added (2024-01-10). Mem0, LightMem, and REMem also answer wrongly; we show the two baselines whose evidence is about the father.} \\
  \bottomrule
\end{extable}
\caption{Example A (whose fact).}
\label{tab:exA}

\vspace{10pt}

\begin{extable}
  \toprule
  \exhead{\textbf{Question} (patient 7, state update): ``If the patient had reduced sertraline to 25\,mg once a day on his own three months ago, what is his current actual dose?'' The records contain no such reduction.} \\
  \exhead{\textbf{Gold}: sertraline 50\,mg every morning.} \\
  \midrule
  \textbf{Embedding RAG, MemOS, LightMem, REMem} \xmark & \textbf{\method} \cmark \\
  \emph{Evidence.} The evidence of Embedding RAG and MemOS contains ``\dots sertraline 50\,mg every morning; I have kept taking it'', but only inside a session text or a narrative memory, with nothing that marks it as the current state.
  &
  \emph{Evidence.} \evgood{``[sertraline $\cdot$ medication chain] [matched $\cdot$ current state] sertraline [active]: started two years ago for anxiety and insomnia; still takes 50\,mg every morning (recorded 2024-03-12)''}; safety alert: ``sertraline with rasagiline: interaction risk; the dose should not be changed without the doctor''. \\
  \emph{Answer} (all four). \evbad{25\,mg}, following the premise of the question.
  &
  \emph{Answer.} The current dose is 50\,mg every morning; the reduction in the question does not appear in the records. \\
  \midrule
  \exwide{\textbf{Why.} The tail of the chain is shown as the current state, so the reader can tell a recorded state from a claim made in the question. Mem0, whose memory says ``currently takes 50\,mg'', also answers correctly.} \\
  \bottomrule
\end{extable}
\caption{Example B (which state is current).}
\label{tab:exB}

\vspace{10pt}

\begin{extable}
  \toprule
  \exhead{\textbf{Question} (patient 4, temporal location): ``In the record of 2024-03-20, what was the patient's blood pressure?''} \\
  \exhead{\textbf{Gold}: 160/95\,mmHg.} \\
  \midrule
  \textbf{Embedding RAG, Mem0, MemOS, REMem} \xmark & \textbf{\method} \cmark \\
  \emph{Evidence.} Embedding RAG retrieves home readings from other days: \evbad{``mostly 145--155/88--92''}.
  &
  \emph{Evidence.} Read at the question's date, the heart-rate, oxygen, and ECG chains return records of that day, e.g., \evgood{``(2024-03-20) heart rate 88, measured at the emergency visit''}. The blood-pressure chain returns its latest reading up to that date, marked as dated 2024-03-02 (home readings, mostly 145--155/88--92). The date channel adds the other records of that day, such as a negative troponin test. The source window of a matched record from that visit reads: \evgood{``\dots I went to the emergency department \dots the blood pressure they measured was 160/95~\dots''}. \\
  \emph{Answer} (Embedding RAG). ``No blood pressure dated 2024-03-20 is in the records.''
  &
  \emph{Answer.} 160/95\,mmHg, measured at the emergency visit on 2024-03-20. \\
  \midrule
  \exwide{\textbf{Why.} The date in the question sets the time at which each chain is read, so the reader gets the measurements of that visit, and every line has its date, so the home readings of 2024-03-02 are not taken for that day's. The source window of a matched record from the same visit gives the reading. LightMem also answers correctly.} \\
  \bottomrule
\end{extable}
\caption{Example C (which date).}
\label{tab:exC}
\end{table*}

\end{document}